\documentclass[pmlr,twocolumn,10pt]{jmlr} 

\usepackage{booktabs}
\usepackage{float}
\usepackage{caption}
\usepackage{siunitx}

\usepackage{array}
\usepackage{makecell} 

\usepackage[switch]{lineno}

\usepackage{listings}
\usepackage{xcolor}
\usepackage{caption}
\usepackage{enumitem}
\DeclareCaptionType{listing}[Listing][List of Listings]

\title[ExECG]{ExECG: An Explainable AI Framework for ECG models}

\author{%
\Name{Jong-Hwan Jang}\footnotemark[1] \Email{jangood1122@medicalai.com}\\
\addr Medical AI Co. Ltd., Seoul, Republic of Korea
\AND
\Name{Yong-yeon Jo}\footnotemark[1] \Email{yy.jo@medicalai.com}\\
\addr Medical AI Co. Ltd., Seoul, Republic of Korea
}

\begin{document}

\maketitle

\begin{abstract}
Deep learning has enabled ECG diagnostic models with strong performance in tasks such as arrhythmia classification and abnormality detection. 
However, accuracy alone is insufficient for clinical deployment because it does not explain why a specific output was produced, limiting justification, error analysis, and trust. 
Although ECG XAI has been extensively investigated and steadily improved, practical pipelines and reporting conventions vary across studies, hindering reuse and reproducibility. 
To address these issues, we present Explainable AI framework for ECG models (\textbf{ExECG}), a Python framework that provides a three-stage pipeline: \texttt{Wrapper} standardizes access across heterogeneous ECG formats and intermediate representations, \texttt{Explainer} unifies diverse XAI methods under a shared execution protocol, and \texttt{Visualizer} supports consistent cross-method comparison within a unified interface. 
We demonstrate end-to-end usage with concise examples and two case studies, highlighting interoperable and reproducible ECG explainability.
\end{abstract}

\paragraph*{Data and Code Availability}
This paper uses the PTB-XL dataset~\citep{ptbxl} available at \url{https://physionet.org/content/ptb-xl/1.0.3/} and the MIMIC-IV ECG dataset~\citep{mimiciv-ecg} available at \url{https://physionet.org/content/mimic-iv-ecg/1.0/}, both accessible on PhysioNet.
\texttt{ExECG} is an open-source framework and available at \url{https://github.com/MAIResearch/ExECG}. All materials and models for needed to reproduce our experiments are included in the repository.

\paragraph*{Institutional Review Board (IRB)}
Since this study only used one publicly available de-identified data, it does not require IRB approval.

\section{Introduction}
\label{sec:intro}

\begin{figure*}[!t]
\centering
\includegraphics[width=\textwidth]{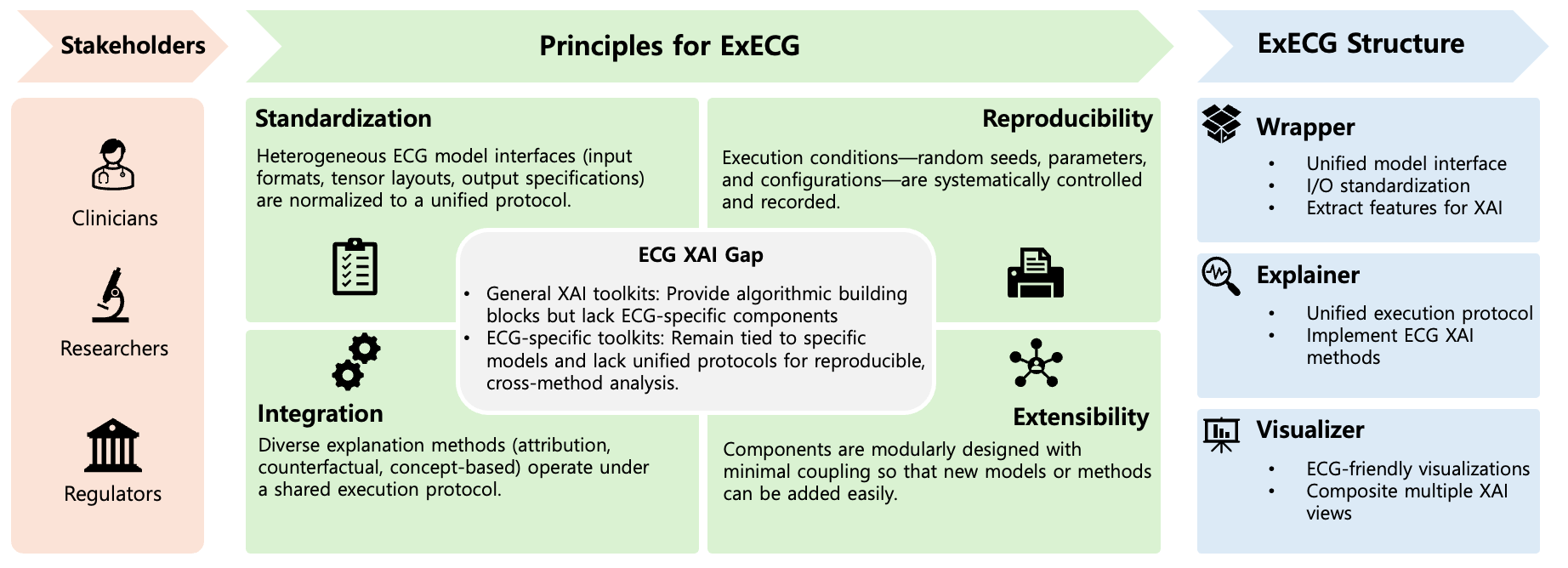}
\caption{\textbf{Design principles and framework overview of \texttt{ExECG}.}
\texttt{ExECG} is developed following four design principles—\textit{Standardization}, \textit{Reproducibility}, \textit{Integration}, and \textit{Extensibility}—motivated by practical needs in clinical and research settings.
The framework operationalizes these principles through three modular components: \texttt{Wrapper}, \texttt{Explainer}, and \texttt{Visualizer}.}

\label{fig:motivation}
\end{figure*}

With the widespread adoption of deep learning, numerous models have been proposed for ECG diagnosis, achieving strong performance on tasks such as arrhythmia classification and cardiac abnormality detection~\citep{wu2025deep,hannun2019cardiologist,ribeiro2020automatic,lim2025ai6lead,shin2025romiae}.
As these models move toward clinical deployment, explainable AI for ECG models (ECG XAI) has become a practical requirement for adoption~\citep{sadeghi2024xaihealthcare,manimaran2025explainable,kelly2019key}.

For ECG XAI to be adopted in real-world practice, it should reflect the distinct needs of different stakeholders.
Clinicians prefer familiar, ECG-native outputs that align with standard ECG reading conventions, since unfamiliar formats are harder to understand and act on~\citep{tonekaboni2019clinicians,bienefeld2023solving,schopp2025clinician}.
Researchers need reproducible pipelines to run, compare, and validate XAI methods~\citep{semmelrock2025reproducibility}.
Regulators expect auditable, consistent procedures that allow results to be verified across evaluations~\citep{kundu2021ai,muehlematter2021approval}.

However, general-purpose XAI frameworks~\citep{lundberg2017unified,kokhlikyan2020captum,schopp2025clinician} still fall short of stakeholder needs in ECG deployment.
They provide generic explanation methods, but offer limited ECG-aware support required for practical workflows; for example, generative counterfactual XAI requires an ECG generator, which is not provided by widely used frameworks.


To alleviate this gap, we introduce \texttt{ExECG} (Explainable AI framework for ECG models), an open-source framework for ECG model explainability built around four design principles: standardization, reproducibility, integration, and extensibility. 
\texttt{ExECG} follows a three-stage pipeline: \texttt{Wrapper} provides a unified interface to heterogeneous ECG models and standardizes access to inputs, outputs, and intermediate representations; \texttt{Explainer} runs diverse XAI methods under a common execution protocol; and \texttt{Visualizer} renders explanations in consistent, clinically aligned formats. 

In addition, \texttt{ExECG} provides ready-to-use materials for ECG XAI, including a StyleGAN-based ECG generator~\citep{jang2025cofe} and concept resources sourced from PhysioNet~\citep{PhysioNet-challenge-2021-1.0.3}.
We demonstrate the end-to-end usage with concise examples and representative use cases, showing how multiple explanation families can be applied and compared within a single, reproducible workflow.

\noindent Our contributions are:
\begin{itemize}
    \item We define four design principles for ECG XAI for the needs of researchers, clinicians, and regulators.
    \item We present \texttt{ExECG}, a three-stage framework following four principles through unified model access, shared execution protocol, and consistent visualization.
    \item We release \texttt{ExECG} to reduce implementation effort for a wide range of ECG studies.
    \item We demonstrate the utility of \texttt{ExECG} by applying diverse XAI methods within a unified workflow.
\end{itemize}


\section{Background}
\label{sec:objective}


\subsection{Existing XAI frameworks and Gaps}
General-purpose XAI frameworks provide algorithmic building blocks but do not fully support ECG explainability workflows.
Captum~\citep{kokhlikyan2020captum} offers gradient-based and concept-based explanation utilities, but applying concept-based methods to ECG requires users to independently curate ECG-relevant concept sets and build experimental infrastructure around them.
SHAP~\citep{lundberg2017unified} focuses on Shapley value-based attribution, making it less suited for workflows that require multiple explanation paradigms.
OmniXAI~\citep{wenzhuo2022-omnixai} offers a unified interface for multiple explanation families, but its support is primarily geared toward tabular and image modalities and does not natively cover multi-channel time-series data such as ECG. Counterfactual-Time-series~\citep{cfts-us-2025} enables counterfactual generation for generic time-series data; however, it likewise does not specifically target or provide ECG-specific components and hard to apply ECG immediately.

An ECG-focused framework exist but offer partial solutions.
ECGxAI~\citep{van2022inherently} was developed primarily for a specific model (FactorECG), and its explanation components are tailored to that model's outputs rather than implemented as reusable, model-agnostic modules.
This tight coupling limits application to other ECG models and hinders extension with new explanation methods. Moreover, existing tools do not present explanations in standardized ECG display formats (e.g., 12-lead grid layouts, calibration markers) familiar to clinicians, limiting integration into clinical reading workflows.

\subsection{Design principles}
\texttt{ExECG} is designed around four principles.

\textbf{Standardization.}
Depending on the study design and data source, the ECG model's input/output specifications can differ. 
\texttt{ExECG} enables researchers to run explanations under a common protocol regardless of the underlying model implementation.
It also reduces the cost of applying explanations to new models.

\textbf{Reproducibility.}
Reproducing explanation requires controlling random seeds, parameters, and execution configurations.
To make this practical, \texttt{ExECG} systematically manages these factors so that explanations can be regenerated under identical conditions.
This supports verification and cross-study comparison.

\textbf{Integration.}
Different explanation methods reveal complementary aspects of model behavior.
Therefore, \texttt{ExECG} consolidates diverse XAI methods under a shared execution protocol with consistent input specifications and output schemas.
This enables systematic multi-method analysis within a single workflow.

\textbf{Extensibility.}
The rapid proliferation of XAI methods has led to fragmented, method-specific pipelines that are difficult to maintain and compare systematically.
To address this, \texttt{ExECG} is designed for plug-and-play extensibility.
It enables new explanation methods to be added or swapped with minimal integration effort.

\section{ExECG: Explainable AI framework for ECG models}
\label{sec:method}

\begin{figure*}[!t]
\centering
\includegraphics[width=\textwidth]{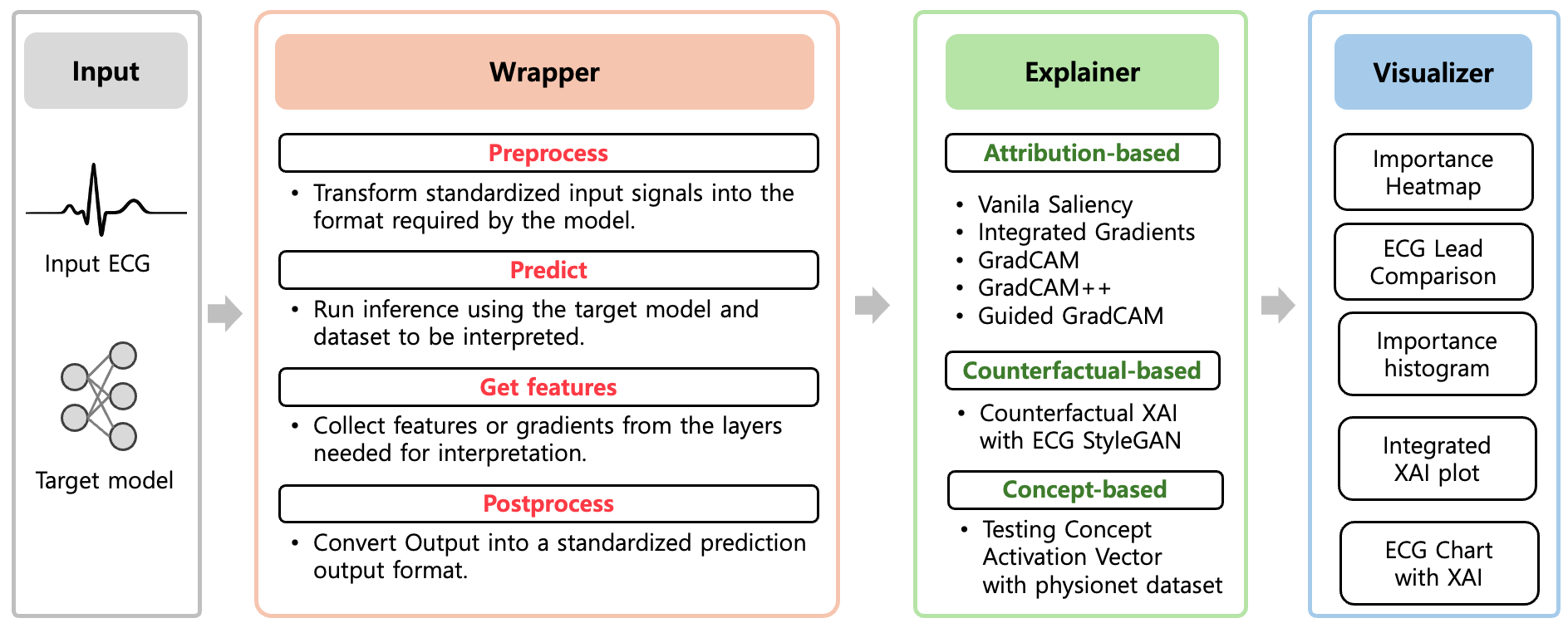}
\caption{Three-stage pipeline of \texttt{ExECG}. 
\texttt{Wrapper} standardizes model I/O and exposes internal signals (e.g., activations and gradients) required by explainers; 
\texttt{Explainer} runs attribution-, counterfactual-, and concept-based XAI methods under a unified interface; 
and \texttt{Visualizer} renders explanation outputs in ECG-aligned plots.}
\label{fig:pipeline}
\end{figure*}

\texttt{ExECG} consists of three stages (Fig.~\ref{fig:pipeline}): \texttt{Wrapper} standardizes different model specifications, \texttt{Explainer} runs XAI methods, and \texttt{Visualizer} renders ECG traces with explanation outputs in consistent formats.
In addition, \texttt{ExECG} provides ECG-specific components: (i) a StyleGAN-based generative model~\citep{jang2025cofe} to support counterfactual explanations, (ii) an ECG concept dataset curated from the PhysioNet/Computing in Cardiology Challenge 2021 dataset~\citep{PhysioNet-challenge-2021-1.0.3} for concept-based methods, and (iii) visualization templates aligned with standard ECG charting conventions.

\subsection{Wrapper}
\texttt{Wrapper} standardizes input/output handling and provides layer-wise access to activations and gradients.

\textbf{Input/Output convention} uses a standardized ECG input shape \texttt{(B, L, T)}, where \texttt{B} is the batch size, \texttt{L} is the number of leads, and \texttt{T} is the signal length in samples.
In contrast, the output format depends on the task type (Table~\ref{tab:io_standardization_example}). 
This convention also fixes key details of the representation (e.g., lead order, sampling rate, and tensor layout) and clarifies how predictions should be reported for each task. 
As a result, XAI methods can run across different models without model-specific parsing or conversion code.

\begin{table}[ht]
\centering
\caption{Standardized Output Specifications. \texttt{B} is the batch size and \texttt{N} is the number of classes or labels.}
\label{tab:io_standardization_example}

\begin{tabular}{l c}
\toprule
\textbf{Task Type} & \textbf{Output Shape} \\
\midrule
Binary classification & \texttt{(B, 2)} \\
Multi-class classification & \texttt{(B, N)} \\
Multi-label classification & \texttt{(B, N)} \\
Regression & \texttt{(B, 1)} \\
\bottomrule
\end{tabular}
\end{table}

\textbf{Feature/gradient access} provides a simple way to retrieve internal signals needed by many XAI methods. 
Users choose one or more target layers, and the interface returns the corresponding activations and, when needed, gradients. 
This makes it easy to apply layer-dependent methods (e.g., gradient-based attribution) in a consistent way across models.

\paragraph{API.}
\texttt{Wrapper} exposes four core API methods that implement the standardized workflow while hiding model-specific details:

\begin{enumerate}[topsep=2pt,itemsep=2pt,partopsep=0pt,parsep=2pt]
  \item \texttt{preprocess}: converts the standardized ECG input into the tensor layout expected by the model.
  \item \texttt{predict}: runs inference under a consistent calling convention and returns raw model outputs.
  \item \texttt{get\_features}: retrieves intermediate representations and, when requested, gradients from user-specified layers.
  \item \texttt{postprocess}: converts raw model outputs (e.g., logits) into the standardized prediction format.
\end{enumerate}

\paragraph{Example.}
Ex~\ref{ex:io_wrapper} shows how the \texttt{Wrapper} handles a difference between the standardized I/O convention and a specific model.
We use a binary classification case where the standardized ECG input is \texttt{(B, L, T)}, but the model expects \texttt{(B, T, L)}.
\texttt{preprocess} reorders the input axes to fit the model, while keeping the XAI pipeline unchanged.
The model outputs logits \texttt{(B, 2)}, and \texttt{postprocess} applies \texttt{softmax} to turn them into class probabilities so the final output matches the standardized convention.

\noindent\begin{minipage}{\columnwidth}
\captionsetup{type=listing, position=top, skip=2pt}
\captionof{listing}{Wrapper usage}
\label{ex:io_wrapper}
\vspace{1mm}
\begin{lstlisting}[language=Python]
wrapper = TorchModelWrapper(
    model,
    preprocess=lambda x: x.transpose(1, 2),
    postprocess=lambda x: torch.softmax(x, dim=-1)
)

output = wrapper.predict(
    inputs,
    output_idx=None, 
    requires_grad=False
)
\end{lstlisting}
\end{minipage}

\subsection{Explainer}

\texttt{Explainer} provides a unified way to execute multiple XAI methods.

\textbf{Unified execution} runs different explanation families through a shared entry point.
Users keep the same standardized ECG input \texttt{(B, L, T)} and a target index, while switching the explainer class to change the explanation method.
This design reduces boilerplate code and makes it easy to compare explanations across methods under the same evaluation protocol.

\textbf{Multi-method support} provides three complementary explanation families.
An attribution explainer produces importance scores aligned with the input waveform (e.g., vanilla saliency map~\citep{simonyan2014deep}, SmoothGrad~\citep{smilkov2017smoothgrad}, Integrated Gradients~\citep{sundararajan2017axiomatic}, Grad-CAM~\citep{selvaraju2017grad}, Grad-CAM++~\citep{chattopadhyay2018gradcampp}, and Guided 
  Grad-CAM~\citep{springenberg2015striving}).
A counterfactual explainer (e.g., StyleGAN based ECG XAI approach ~\citep{jang2025cofe}) generates modified ECG signals and the corresponding model responses toward a desired target value~\citep{jang2025novel}.
A concept-based explainer (e.g., TCAV~\citep{kim2018interpretability}) computes concept sensitivity scores at user-specified layers, enabling global-level explanation beyond local explanation for ECG models ~\citep{lee2025transparent}. 

\paragraph{API.}
\texttt{Explainer} classes follow a common API pattern:
\begin{enumerate}[topsep=2pt,itemsep=2pt,partopsep=0pt,parsep=2pt]
  \item \texttt{explain}: runs the selected explanation method for \texttt{inputs} and a given \texttt{target}.
  (e.g., a dictionary with \texttt{results} for attribution, a tuple for counterfactual signals, or a score table for TCAV).
\end{enumerate}

\begingroup
\setlength{\parskip}{2pt}
\paragraph{Examples.}
In Ex~\ref{ex:explainer_three_examples}, we provide three representative examples covering attribution-, counterfactual-, and concept-based explainers.
All examples follow the same calling pattern via the \texttt{explain} function, while method-specific arguments determine the explanation type and configuration.

\textbf{Case 1:} This computes a layer-specific attribution map.
The user specifies the layer to explain (e.g., \texttt{target\_layer\_name="conv3"}) and selects the Grad-CAM variant with \texttt{method="gradcam"} for \texttt{target=0}.

\textbf{Case 2:} This generates a counterfactual ECG that reaches the desired target value.
A StyleGAN-based counterfactual explainer is instantiated with \texttt{input\_sampling\_rate=250} and synthesizes an alternative ECG that shifts the model output toward \texttt{target\_value=1.0} for \texttt{target=0}.

\textbf{Case 3:} This evaluates concept sensitivity at a designated layer.
Concepts are sourced from the open dataset~\citep{PhysioNet-challenge-2021-1.0.3}, and TCAV compares \texttt{"atrial fibrillation"} versus \texttt{"sinus rhythm"} for \texttt{target=1} using \texttt{model\_layers\_list=["conv3"]}.
\endgroup

\noindent\begin{minipage}{\columnwidth}
\captionsetup{type=listing, position=top}
\captionof{listing}{Three \texttt{Explainer} usages}
\label{ex:explainer_three_examples}
\vspace{1mm}
\begin{lstlisting}[language=Python]
# Case 1: Attribution (Grad-CAM)
explainer = GradCAM(wrapper)
result = explainer.explain(
    ecg_tensor,
    target=0,
    target_layer_name="conv3",
    method="gradcam"
)

# Case 2: Counterfactual (StyleGAN-based)
explainer = StyleGANCF(
    model=wrapper,
    input_sampling_rate=250
)
result = explainer.explain(
    ecg_tensor,
    target=0,
    target_value=1.0
)

# Case 3: Concept-based (TCAV)
explainer = TCAV(
    model=wrapper,
    model_layers_list=["conv3"],
    input_sampling_rate=250,
    input_duration=10,
    data_name="physionet2021",
    target_concepts=["atrial fibrillation", "sinus rhythm"]
)
result = explainer.explain(ecg_data, target=1)
\end{lstlisting}
\end{minipage}


\subsection{Visualizer}
\texttt{Visualizer} provides ECG-aligned plots for each XAI family and a compact integrated view that combines multiple explanations.

\textbf{Method-specific rendering} provides dedicated visualizations for attribution-, counterfactual-, concept-based explanations, and an integrated ECG view.
All plots follow the standardized ECG representation and are aligned to the lead and time axes.

For attribution-based explanations, importance scores are displayed as time-binned heatmaps aligned to the waveform.
For counterfactual-based ones, the original and counterfactual signals are overlaid on the same lead to highlight the changes.
For concept-based ones (TCAV), results are summarized as layer-wise concept sensitivity scores with optional confidence intervals.
Finally, an integrated ECG chart combines the waveform with multiple XAI layers (e.g., attribution overlays and counterfactual traces) in a single view.

\paragraph{API.}
\texttt{Visualizer} provides four plotting functions that cover common ECG XAI use cases:
\begin{enumerate}[topsep=2pt,itemsep=2pt,partopsep=0pt,parsep=2pt]
  \item \texttt{plot\_attribution}: visualizes lead-wise attribution aligned to the ECG waveform (optionally time-binned).
  \item \texttt{plot\_counterfactual\_overlay}: overlays original and counterfactual signals for a selected lead with prediction summaries.
  \item \texttt{plot\_tcav\_scores} (and \texttt{plot\_tcav\_ci}): visualizes concept sensitivity scores and their confidence intervals at specified layers.
  \item \texttt{plot\_ecg\_chart}: renders an integrated ECG chart with optional counterfactual overlays and attribution heatmaps.
\end{enumerate}

\begin{figure*}[t]
\centering
\fbox{\includegraphics[width=\textwidth]{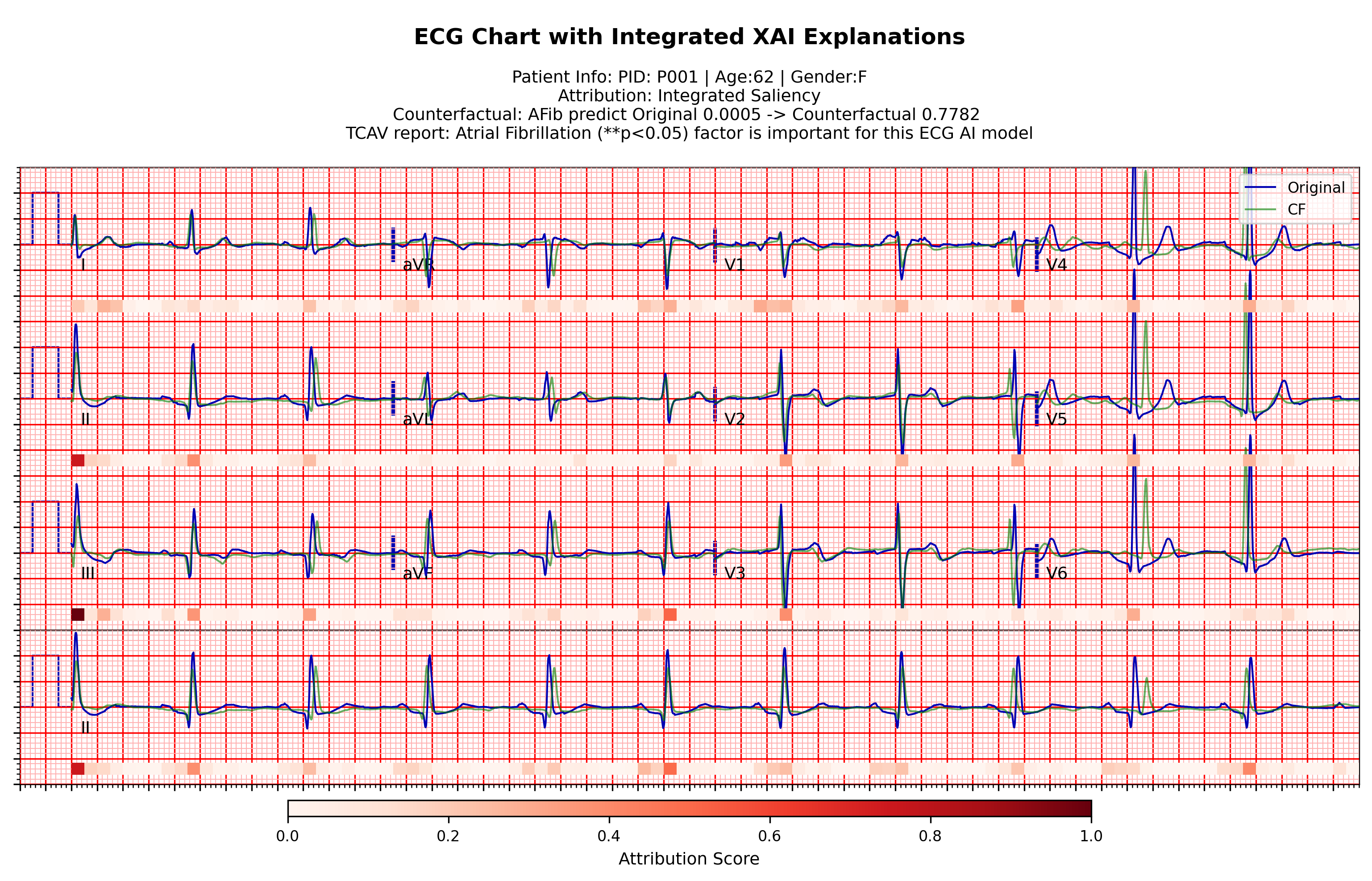}}
\caption{\textbf{Integrated XAI visualization in standard 12-lead ECG format.}
\texttt{ExECG} renders explanations on the clinical 4$\times$3 grid layout familiar to practitioners.
The chart overlays the original ECG (blue) with counterfactual (green), displays attribution importance as background shading, and reports TCAV-derived concept significance, enabling multi-method interpretation within a single view.}
\label{fig:ecg_chart}
\end{figure*}

\begingroup
\setlength{\parskip}{2pt}
\paragraph{Examples.}
Ex~\ref{ex:viz_examples} summarizes four typical visualization calls.
All cases follow a consistent plotting interface, while the inputs and optional arguments determine the visualization type and overlays.

\textbf{Case 1:} This visualizes an attribution heatmap with optional time binning to improve readability.
The user sets \texttt{bin\_size=25} to aggregate attribution scores over time and reduce high-frequency noise (e.g., Fig.~\ref{fig:attr_afib}).

\textbf{Case 2:} This overlays an original ECG and a counterfactual ECG on a selected lead.
The user selects \texttt{lead\_idx=1} and annotates the target-class probabilities for the original and counterfactual signals (e.g., Fig.~\ref{fig:cf_afib}).

\textbf{Case 3:} This reports concept sensitivity scores and their uncertainty estimates for a specified layer.
The user plots TCAV scores and confidence intervals for \texttt{"conv3"} using \texttt{plot\_tcav\_scores} and \texttt{plot\_tcav\_ci} (e.g., Fig.~\ref{fig:tcav_afib}).

\textbf{Case 4:} This renders an integrated ECG chart that combines multiple XAI overlays in a single view.
The user enables calibration markers and overlays counterfactual traces and attribution heatmaps (e.g., \texttt{show\_calibration=True}, \texttt{cf\_ecg}, and \texttt{attribution\_bin\_size=25}) (e.g., Fig.~\ref{fig:ecg_chart}).
\endgroup

\noindent\begin{minipage}{\columnwidth}
\captionsetup{type=listing, position=top, skip=2pt}
\captionof{listing}{Four \texttt{Visualizer} usages}
\label{ex:viz_examples}
\vspace{1mm}
\begin{lstlisting}[language=Python]
# Case 1: Attribution heatmap (time-binned)
plot_attribution(
    ecg_data,
    importance_score,
    bin_size=25
)

# Case 2: Counterfactual overlay on a selected lead
plot_counterfactual_overlay(
    original_ecg,
    counterfactual_ecg,
    lead_idx=1,
    original_prob=original_prob,
    counterfactual_prob=counterfactual_prob
)

# Case 3: TCAV sensitivity scores and confidence intervals
plot_tcav_scores(tcav_results)
plot_tcav_ci(tcav_results, layers=["conv3"])

# Case 4: Integrated ECG chart with XAI overlays
plot_ecg_chart(
    ecg,
    sample_rate=250,
    title="ECG Chart with integrated XAI",
    columns=4,
    show_calibration=True,
    cf_ecg=cf_ecg,
    cf_color="green",
    cf_alpha=0.6,
    attribution=attrituion_score,
    attribution_bin_size=25,
    attribution_cmap="viridis"
)
\end{lstlisting}
\end{minipage}






\section{Use Cases}
\label{sec:case}

We demonstrate \texttt{ExECG} on an atrial fibrillation (AF) detection as the binary classification task~\footnote{We also apply \texttt{ExECG} to potassium-level estimation as a regression task; details are provided in Appendix~\ref{apd:first}.}.
We used a 1D ResNet backbone~\citep{he2016deep} with a lightweight binary classification head that outputs class probabilities.
The model was trained on 10-second, 12-lead ECG segments sampled at 250\,Hz, using the PTB-XL dataset~\citep{ptbxl} and the MIMIC-IV ECG dataset~\citep{mimiciv-ecg}.
The model training procedures follow \citep{jang2025novel} protocol.

\subsection{Attribution-based Use Case}
\begin{figure*}[h]
\centering
\includegraphics[width=\textwidth]{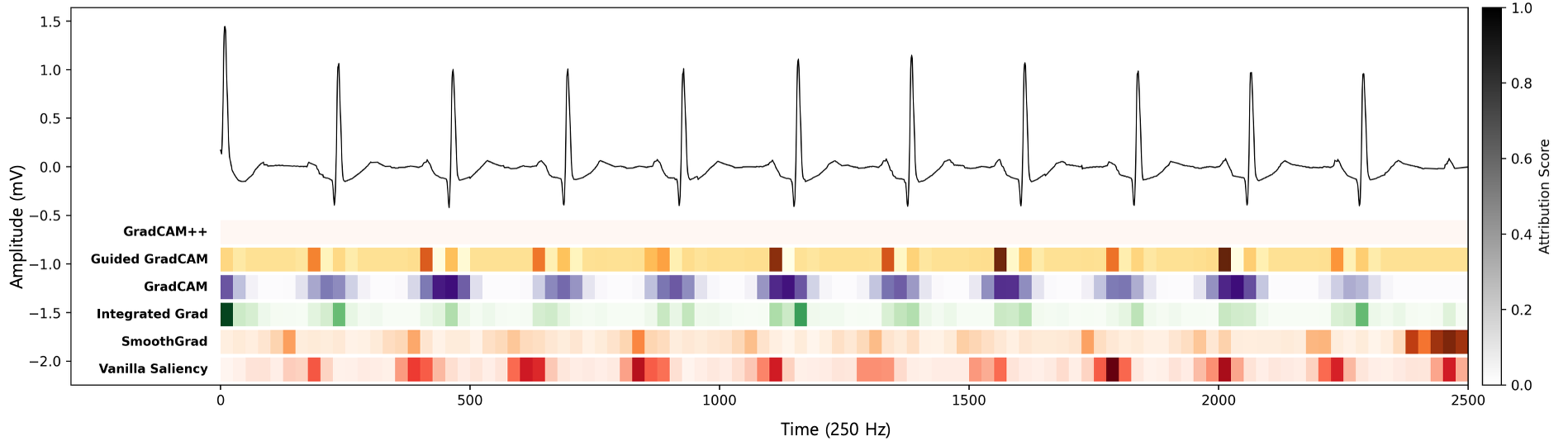}
\caption{\textbf{An Example of attribution methods for AF classification.}
Six attribution methods are applied to the same ECG sample, with importance scores displayed as aligned heatmaps below the waveform. Most methods consistently highlight the P-wave region, suggesting the model relies on atrial activity for AF detection.}
\label{fig:attr_afib}
\end{figure*}

Fig.~\ref{fig:attr_afib} illustrates multiple attribution methods for a single lead.
Guided Grad-CAM, Grad-CAM, SmoothGrad, and the vanilla saliency map consistently highlight the same region of the ECG waveform.
This region corresponds to the P-wave segment, which is often absent or markedly attenuated in AF~\citep{hindricks2021esc}.
The agreement across explainers therefore suggests that the classifier relies on P-wave.
Unlike the four methods above, Integrated Gradients and GradCAM++ highlight different regions of the signal, yielding less consistent attribution patterns.


\subsection{Counterfactual-based Use Case}
\begin{figure*}[h]
\centering
\includegraphics[width=\textwidth]{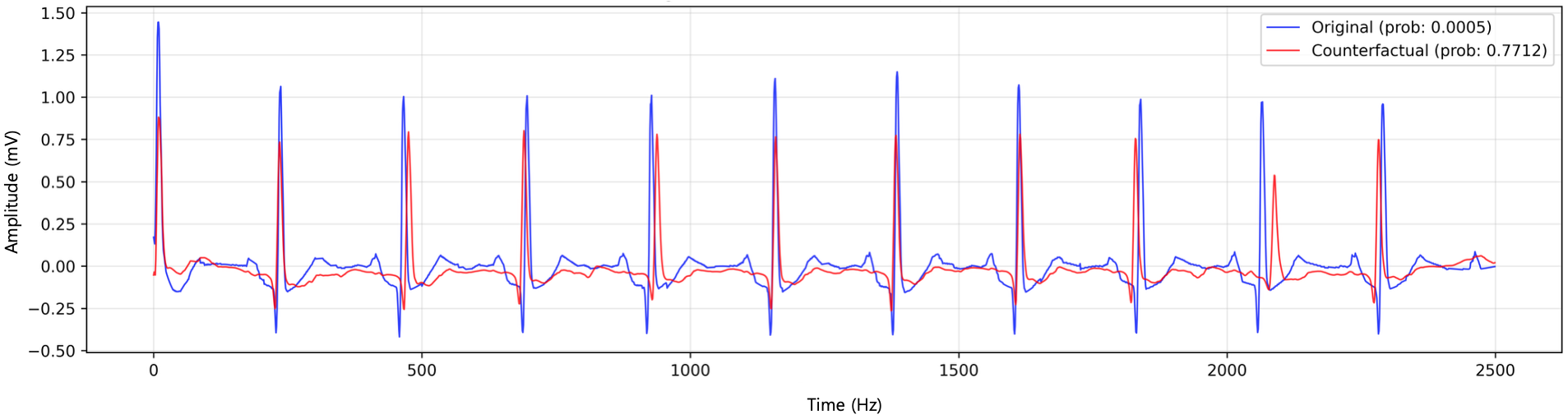}
\caption{\textbf{An Example of counterfactual explanation for AF classification.}
The original ECG (blue, AF probability 0.0005) is overlaid with its counterfactual (red, AF probability 0.7712).The counterfactual shows P-wave attenuation and irregular RR intervals, which are characteristic patterns of AF.}
\label{fig:cf_afib}
\end{figure*}

Fig.~\ref{fig:cf_afib} overlays the original ECG (blue) and the counterfactual ECG (red).
The counterfactual one increases the AF probability from 0.0005 to 0.7712 by introducing clinically relevant waveform changes that shift the model output toward the target class.
The counterfactual may also introduce irregular waveform morphology, leading to increased peak-to-peak variability that is not directly attributable to AF~\citep{lip2016atrial}, which can complicate interpretation.


\paragraph{Summary.}
Across these use cases, \texttt{ExECG} enables cross-method inspection of model behavior under a single, consistent workflow.
Attribution results provide instance-level evidence of salient waveform cues, while counterfactual analysis exposes how plausible signal modifications can shift the model output toward AF.
Together, these complementary views help validate whether the model relies on clinically meaningful patterns and reveal cases where explanations become inconsistent or difficult to interpret.

\subsection{Concept-based Use Case}
\begin{figure*}[h]
\centering
\includegraphics[height=6cm,width=\textwidth]{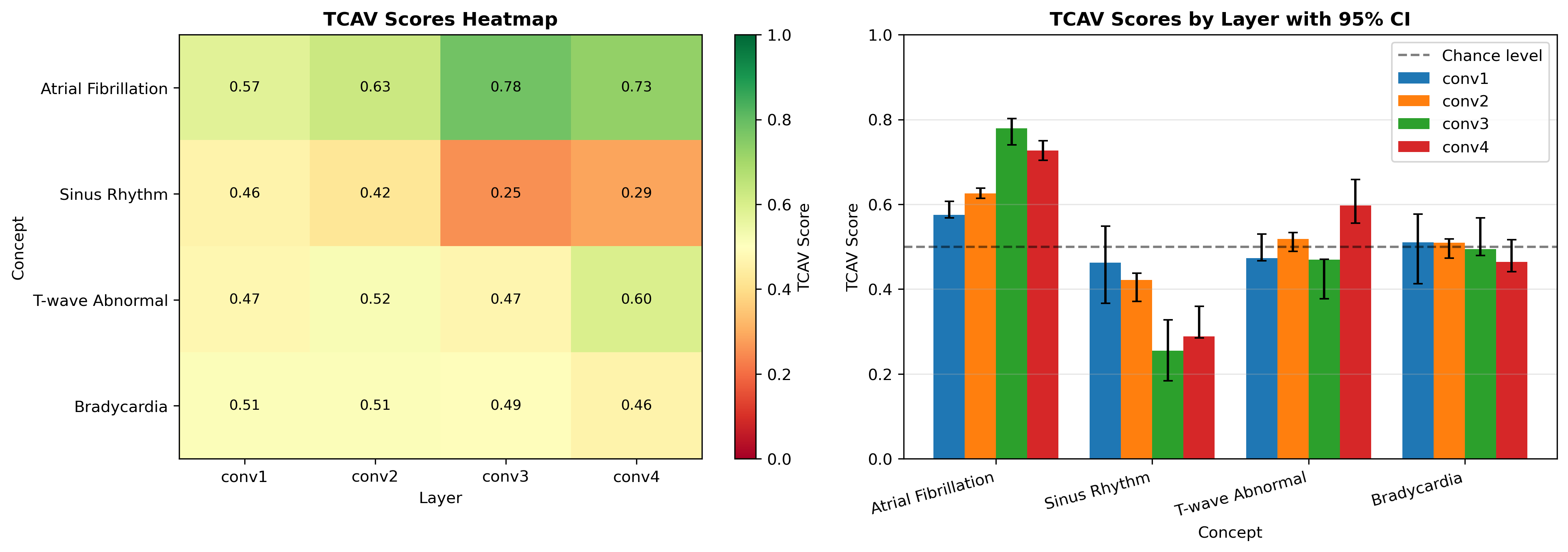}
\caption{\textbf{TCAV Analysis for AF Classification.}
Left: Heatmap of TCAV scores across four clinical concepts and network layers. Right: TCAV scores with 95\% confidence intervals.
The \textit{atrial fibrillation} concept shows scores consistently above 0.5 (chance level) across all layers, indicating stable reliance on AF-related features, while other concepts remain near or below chance.}
\label{fig:tcav_afib}
\end{figure*}

Fig.~\ref{fig:tcav_afib} summarizes TCAV results for four concepts related to cardiac rhythm and conditions: {atrial fibrillation}, {sinus rhythm}, {T-wave abnormal}, and {bradycardia}.
The {atrial fibrillation} concept exhibits consistently high TCAV scores across layers, with confidence intervals remaining well above 0.5, suggesting that the model repeatedly leverages AF-related evidence.
In contrast, {sinus rhythm} shows comparatively lower scores, while the remaining concepts stay close to 0.5, implying limited influence on the model prediction.
Overall, TCAV trends provide semantic evidence that helps contextualize the model decision.


\section{Conclusion}
\label{sec:conclusion}
We presented \texttt{ExECG}, an open-source Python framework for explaining deep learning models for electrocardiography.
\texttt{ExECG} is organized as a modular three-stage pipeline: \texttt{Wrapper} standardizes model access and exposes intermediate representations required by XAI, \texttt{Explainer} executes multiple explanation families through a consistent interface, and \texttt{Visualizer} renders results in ECG-aligned formats for transparent inspection and reporting.
Through an AF classification case study, we demonstrated how attribution-, counterfactual-, and concept-based explainers provide complementary views of model behavior under a consistent workflow.
By publicly releasing the codebase, configurations, and documentation, \texttt{ExECG} lowers the practical barrier to reproducible ECG XAI and enables straightforward extension to new models and explanation methods.
We expect \texttt{ExECG} to support systematic evaluation of explainability in ECG AI and facilitate progress toward clinically relevant and trustworthy deployment.
We hope this framework serves as a foundation for advancing ECG XAI research.
Finally, our current evaluation is limited to a single use case, and broader validation across tasks, datasets, and architectures remains future work.

\clearpage
\twocolumn

\raggedbottom
{\small
\bibliography{chil-sample}
}

\clearpage
\appendix
\section{Usecase: Potassium level estimation}
\label{apd:first}
\subsection{Model Development}
Both the AF classification model and the potassium regression model follow the same training protocol as described in GCX~\citep{jang2025novel}.
We use a 1D ResNet architecture with a task-specific prediction head.
The input is a 10-second, 12-lead ECG segment sampled at 250\,Hz, resulting in a tensor of shape $(12 \times 2500)$.

\textbf{Potassium Regression Model.}
The potassium model is trained on MIMIC-IV ECG~\citep{mimiciv-ecg} for continuous potassium level estimation.
The model outputs a single scalar value representing the predicted serum potassium level (mEq/L).
Training follows the same protocol as the AF model, with the loss function changed from cross-entropy to mean squared error (MSE) for the regression task.

Both models are trained until convergence and the best checkpoint is selected based on validation performance.
Full implementation details, including exact hyperparameters and training curves, are available in the GCX repository.

\subsection{Clinical background}
Potassium changes produce three key ECG patterns: \textbf{peaked T-waves}, \textbf{QRS widening}, and \textbf{prolonged PR interval}~\citep{mattu2000electrocardiographic,diercks2004electrocardiographic}.
We use these patterns to assess explanation validity.

\subsection{Attribution-based}
\begin{figure*}[!t]
\centering
\includegraphics[width=\textwidth]{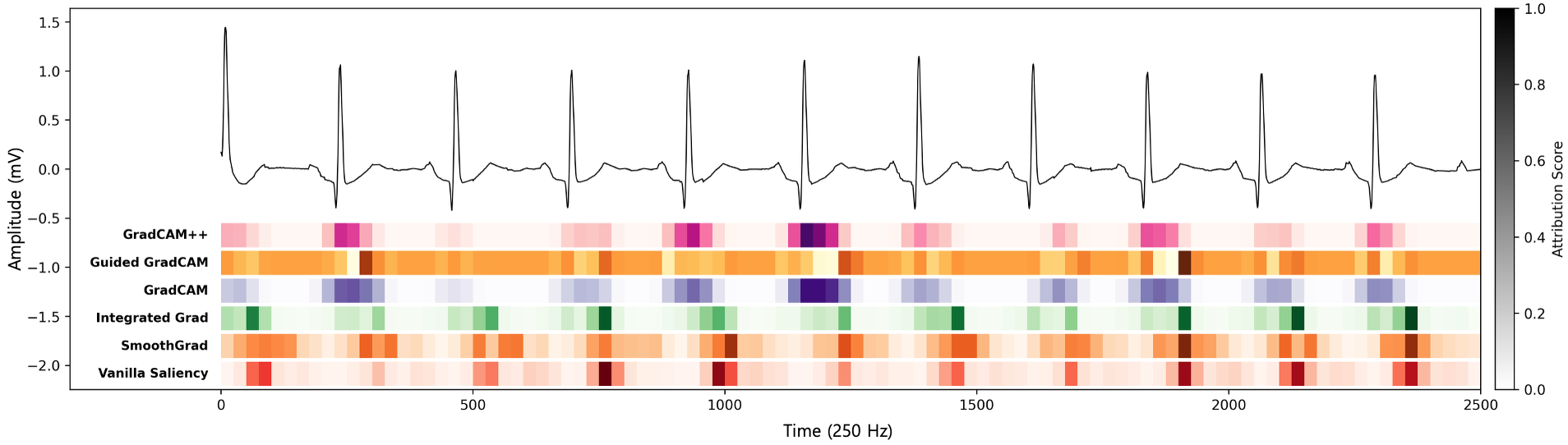}
\caption{\textbf{Comparison of Attribution Methods for Potassium Level Estimation.}
Attribution heatmaps for six XAI methods applied to a potassium regression model.
Integrated Gradients, SmoothGrad, and Vanilla Saliency consistently emphasize T-wave regions, while Grad-CAM variants assign broader importance across the QRS-T segment.}
\label{fig:attr_pota}
\end{figure*}

Fig.~\ref{fig:attr_pota} compares attribution methods for the potassium regression model.
Grad-CAM variants assign importance broadly across the QRS complex and T-wave, suggesting the model relies on the overall depolarization-repolarization segment.
Guided Grad-CAM highlights many regions and appears noisier.
In contrast, Integrated Gradients, SmoothGrad, and Vanilla Saliency consistently emphasize the T-wave region, indicating T-wave morphology is dominant.
Overall, attribution identifies ST-T evidence but does not show \emph{how} the signal should change to alter predicted potassium values.

\subsection{Counterfactual-based}
\begin{figure*}[t]
\centering
\includegraphics[width=\textwidth]{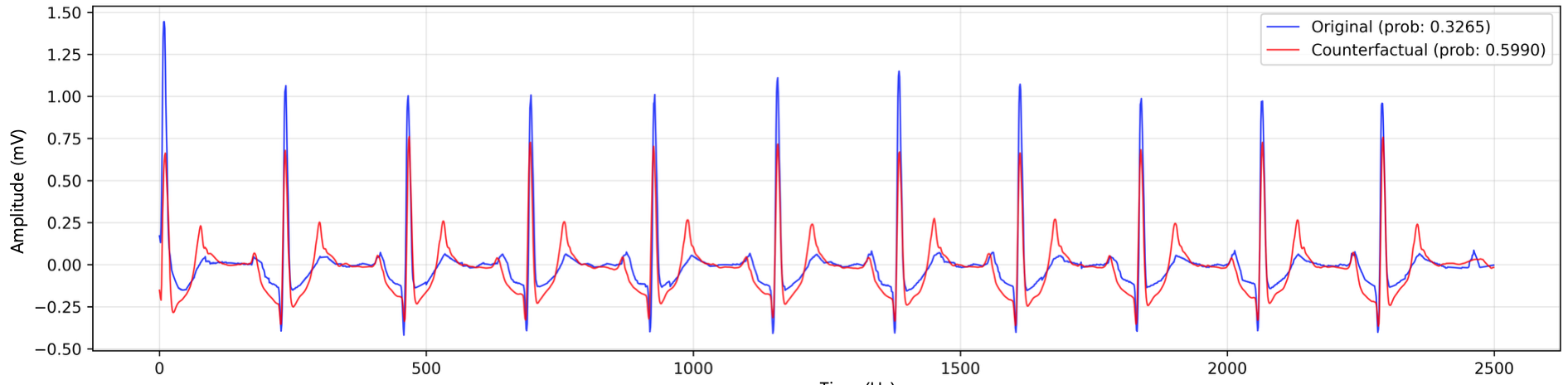}
\caption{\textbf{Counterfactual Explanation for Potassium Level Estimation.}
The original ECG (blue, predicted value 0.33) is overlaid with its counterfactual (red, predicted value 0.60).
The counterfactual shows peaked T-waves, widened QRS complexes, and prolonged PR intervals—characteristic ECG changes associated with hyperkalemia.}
\label{fig:cf_pota}
\end{figure*}

Fig.~\ref{fig:cf_pota} overlays the original ECG (blue) and counterfactual (red).
The model prediction increases from 0.005 to 0.5395.
The counterfactual shows potassium-related changes: the T-wave becomes more peaked, the QRS complex widens, and the P-wave shifts to \textbf{prolong the PR interval}.
Notably, this PR-interval change is not evident in attribution maps, showing how counterfactuals reveal clinically meaningful alterations that attribution misses.

\subsection{Concept-based}
\begin{figure*}[!t]
\centering
\includegraphics[width=\textwidth]{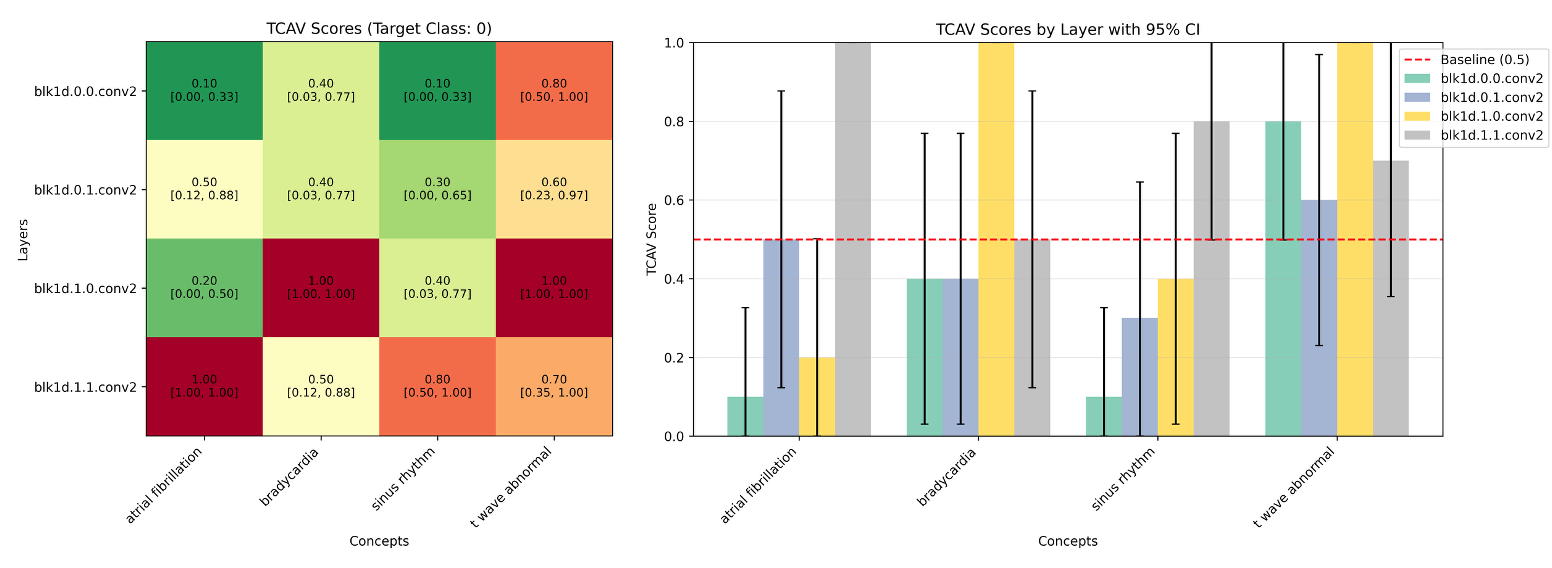}
\caption{\textbf{TCAV Analysis for Potassium Level Estimation.}
Left: Heatmap of TCAV scores for four clinical concepts across network layers. Right: TCAV scores with 95\% confidence intervals.
The \textit{T-wave abnormal} concept exhibits scores significantly above 0.5, indicating that the model relies on T-wave morphology for potassium prediction.}
\label{fig:tcav_pota}
\end{figure*}

Fig.~\ref{fig:tcav_pota} shows TCAV results for the same four concepts.
\textit{T-wave abnormal} shows consistently high TCAV scores across layers, with confidence intervals significantly above 0.5.
This indicates the model relies on T-wave abnormality for potassium prediction.
Other concepts remain near 0.5, indicating weaker influence.

\end{document}